\documentclass[conference]{IEEEtran}
\IEEEoverridecommandlockouts
\usepackage{cite}
\usepackage{amsmath,amssymb,amsfonts}
\usepackage{algorithmic}
\usepackage{graphicx}
\usepackage{textcomp}
\usepackage{subcaption}
\usepackage{xcolor}
\def\BibTeX{{\rm B\kern-.05em{\sc i\kern-.025em b}\kern-.08em
    T\kern-.1667em\lower.7ex\hbox{E}\kern-.125emX}}
\begin{document}

\title{Local Path Planning
and Obstacle
Avoidance \\ for an
Omnicopter Platform
}

\author{
Mikołaj Heliński$^{1}$,
Spilios Theodoulis$^{1}$,
Mahmoud Hamandi$^{2}$,
Abdullah Mohamed Ali$^{3}$,\\
Anthony Tzes$^{3}$,
Marija Popovi\'c$^{1}$

\thanks{$^{1}$M. Heliński, S. Theodoulis, and M. Popovi\'c are with the Faculty of Aerospace Engineering, Delft University of Technology, Delft, The Netherlands. {\tt\small helinskimikolaj@gmail.com}, {\tt\small \{S.Theodoulis, M.Popovic\}@tudelft.nl}}

\thanks{$^{2}$M. Hamandi is with Mohamed Bin Zayed University of Artificial Intelligence, Abu Dhabi, UAE. {\tt\small mahmoud.hamandi@mbzuai.ac.ae}}

\thanks{$^{3}$A. Mohamed Ali and A. Tzes are with the Center for Artificial Intelligence and Robotics, New York University Abu Dhabi, Abu Dhabi, UAE. {\tt\small \{abdullah.ali, anthony.tzes\}@nyu.edu}}
}

\maketitle

\begin{abstract}

Autonomous unmanned aerial vehicles (UAVs) increasingly operate in cluttered environments where global planners such as RRT* are not directly deployable at control rates. This paper presents a real-time local planning and obstacle avoidance module for an omnidirectional multirotor (omnicopter) by extending the Dynamic Window Approach to six degrees of freedom (6D-DWA). Our method achieves real-time feasibility through (i) local-map voxelisation, (ii) a compact sphere-based approximation of the vehicle geometry, and (iii) adaptive velocity sampling in the 6D search space. To improve reactivity to unknown obstacles, we introduce a context-aware ``Agile Mode” that adjusts scoring weights online to trade-off between goal progress, clearance, and heading/facing constraints during evasive manoeuvres. We evaluate our approach in simulation across computational stress tests, dense-waypoint path tracking, and static/unknown obstacle scenarios. Our planner runs consistently within a $0.2$\,s control loop, tracks waypoint-dense global paths with $<0.1$\,m average cross-track error and $\sim13^\circ$ average heading error, and avoids collisions in static environments. For unknown obstacle avoidance, Agile Mode achieves 79.3\% success for an off-centre obstacle and 41.4\% for a centred obstacle, highlighting both the effectiveness of adaptive weighting and remaining limitations in highly constrained geometries.
\end{abstract}


\section{Introduction}
\label{sec:introduction}
Recent years have seen rapid development in autonomous multicopter platforms~\cite{UAVpath}. The ability of multicopter UAVs to perform precise hovering and operate at low horizontal velocities makes them particularly useful for tasks such as search and rescue, delivery, and surveillance~\cite{ClassificationUAV}. Beyond conventional underactuated vehicles, omnidirectional multirotor, e.g. fully actuated octocopter designs, provide independent control of translational and rotational motion, enabling increased manoeuvrability and improved attitude control authority~\cite{Fully, full_act}. This omnidirectionality expands the range of potential applications, including interaction-rich flight and aerial manipulation~\cite{aerialmanipulation}.

Autonomous navigation in cluttered environments is typically organised as a hierarchy that combines global planning with local execution~\cite{algcomp}. Global planners can generate collision-free routes towards a goal, while local planners optimise trajectory execution and perform real-time replanning when sensing uncertainty or unforeseen obstacles are encountered~\cite{local1}. A broad range of planning methods exists for 3D traversal, exhibiting trade-offs between computational complexity and performance~\cite{meta}. In dynamic environments, obstacle detection and avoidance methods are often only loosely coupled to the planning pipeline, which can reduce robustness when operating under real-time constraints~\cite{local1, pathsurvey}. While hybrid pipelines may mitigate weaknesses of individual components, they frequently increase computational demand, which is particularly limiting for omnidirectional platforms due to the high dimensionality of a 6D search space.


This paper addresses the need for a collision-aware, real-time feasible local planner for an omnidirectional multirotor operating within a broader autonomy stack. We build on an existing omnidirectional octocopter platform~\cite{nyuad_drone} and a rapidly exploring random trees (RRT)*-based global planner that outputs 6D waypoints~\cite{nyuad_RRT*}.  RRT* is not directly suitable for online local replanning at control rates~\cite{rrtstar}. Also, executing global waypoints without reactive avoidance can lead to collisions with unmapped or dynamic obstacles, including wall-induced aerodynamic suction effects that can draw the UAV toward nearby surfaces \cite{omniiros}. To tackle this issue, we propose a computationally efficient 6D extension of the Dynamic Window Approach (6D-DWA) that samples feasible linear and angular velocities and selects commands using a weighted objective while enforcing collision constraints. Our method enables real-time performance within a $5$\,Hz replanning through (i) voxelisation of the local map for efficient collision queries~\cite{voxel}, (ii) a compact approximation of the complex vehicle geometry, and (iii) adaptive sampling strategies that focus computation on promising regions of the 6D velocity space. Further, we incorporate a context-aware mode that adjusts scoring weights online to improve reactive avoidance behaviour when unforeseen obstacles are encountered.

The main contributions of this paper are:
\begin{enumerate}
    \item a 6D extension of DWA suitable for omnidirectional multirotors, including feasibility constraints over both linear and angular velocity components;
    \item a set of practical computational optimizations, i.e. voxel map representation, geometry approximation, adaptive sampling, enabling operation within a $5$\,Hz replanning cycle;
    \item a simulation-based evaluation in Gazebo/ROS~\cite{gazebo, ros} assessing runtime, waypoint-dense path tracking, and obstacle avoidance in static and previously unmapped scenarios.
\end{enumerate}


\section{Proposed 6D-DWA Architecture}
\label{local}

Our local planner must support real-time replanning at control rates. To this end, we build on the Dynamic Window Approach (DWA)~\cite{dwa1}, which samples velocity commands that are physically reachable from the current state and forward-simulates the resulting motion over a short horizon. Candidate trajectories are rejected if they violate collision constraints, and the remaining set is ranked using a weighted objective that encodes the desired behaviour (e.g. goal progress and safety). Restricting evaluation to dynamically feasible commands both reduces the search space and yields predictable runtime, enabling behavioural tuning without modifying the core planner. DWA is most commonly used in planar settings, since naively increasing the search dimensionality quickly becomes computationally prohibitive~\cite{3d-dwa}.

For an omnidirectional multirotor, however, the local decision space is inherently six-dimensional, comprising the 3D linear and angular velocities denoted respectively $(V_x, V_y, V_z, \omega_x, \omega_y, \omega_z)$. The key challenge is therefore to retain DWA’s reactivity while making a 6D search tractable in real time: sampling density must be limited, and collision checking must be implemented efficiently. In this work we target a $5$\,Hz replanning cycle ($0.2$\,s per iteration) and design the sampling and collision evaluation accordingly. Finally, because DWA optimises only over a short horizon and returns a single best command per cycle, it is guided by a higher-level plan: global waypoints provide long-horizon intent, while the proposed 6D-DWA remains responsible for local feasibility and reactive obstacle avoidance.

\subsection{6D Velocity Sampling}
\label{subsec:velocity_sampling}
Our 6D-DWA evaluates sampled velocity commands within the Dynamic Window, defined as the set of linear and angular velocities that are reachable from the current omnicopter state within a single time step ($\Delta t$). This set is bounded by hardware-imposed limits on velocity and acceleration, as well as the vehicle's current linear and angular velocities.

A significant challenge in extending DWA to 6D-DWA is the exponential growth of the search space compared to the planar case. Since each sampled velocity must be forward-simulated and scored, the number of samples directly determines the computational load per planning cycle. To maintain real-time performance, rather than distributing samples uniformly over the entire Dynamic Window, we introduce an Adaptive Sampling strategy, that focuses sampling on areas with a higher likelihood of high-scoring velocity inputs.

Sampling is divided into three stages: Exploration ($e_r$) for unbiased space coverage, Focused-Search ($f_r$) for sampling around the previous best command with a standard deviation $\sigma$, and Boundary-Search ($b_r$) to evaluate performance at the Dynamic Window limits. These ratios define the proportion of the total sample count allocated to each stage ($e_r + f_r + b_r = 1.0$).

\subsection{Static Obstacle Collision Avoidance}
\label{subsec:static_collision_detection}
Throughout this paper, the proposed path planning approach focuses on collision avoidance between the omnicopter platform presented in~\cite{nyuad_drone} and its surrounding environment. The planner generates trajectories that simultaneously avoid collisions with both moving obstacles (addressed in a later section) and static obstacles. Static obstacles are assumed to be pre-mapped, with their geometry stored in an STL file. To enable real-time performance, static collision detection is implemented using a combination of vehicle structure simplification, environment voxelisation, and dynamic local map generation.    

\begin{figure}[b]
    \centering
    \includegraphics[width=0.95\columnwidth]{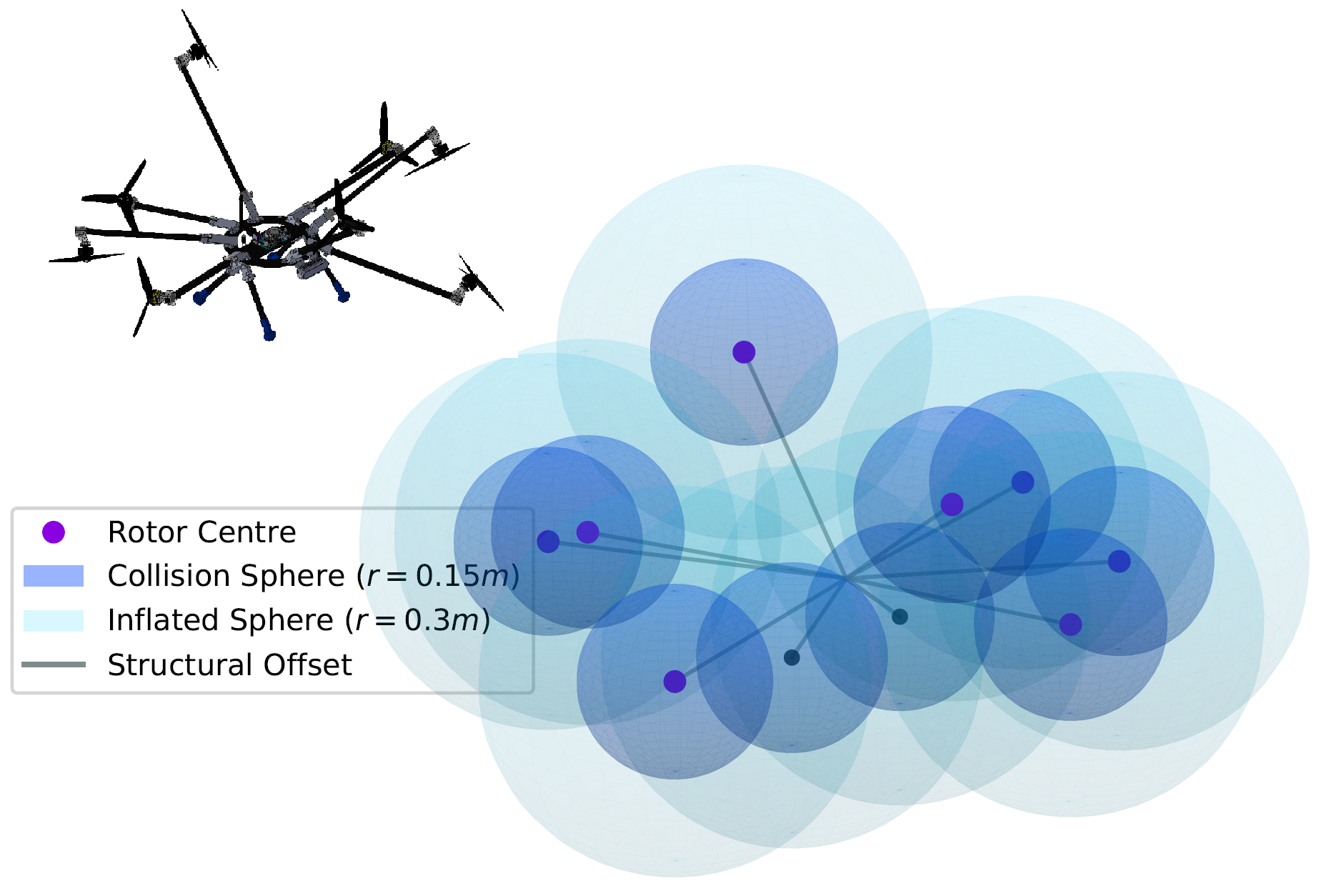}
    \caption{The 10-sphere approximation covering the UAV's structure on the right, and the actual UAV on the left.}
    \label{fig:spheres}
\end{figure}

The omnicopter’s complex geometry makes exact mesh-based collision checking computationally prohibitive for real-time use. Therefore, the collision volume of the platform is approximated using ten spheres covering the body of the vehicle, as presented in Fig.~\ref{fig:spheres}. This representation enables efficient distance-based collision checking that is invariant to vehicle orientation. Each sphere radius is artificially inflated by a safety margin to ensure robust avoidance despite geometric approximations or tracking errors; the corresponding inflated radius is referred to as $r_{\text{inf}}$.

To represent the static environment, the STL mesh undergoes recursive densification by bisecting triangle edges until the maximum spacing between adjacent surface points falls below a predefined threshold. This guarantees bounded surface resolution and prevents missed collisions due to coarse triangulation. The resulting point set is then voxelised: all surface points contained within a voxel are replaced by a single centroid. This reduces the number of obstacle points while preserving geometric coverage~\cite{Occupancygrid}. The occupied voxels are stored in a set $\mathcal{O}_{\text{obs}}$, where each entry $\mathbf{o}_k \in \mathbb{R}^3 \,\, \forall k \in \{1,N\}$ represents the coordinate of an occupied voxel.

Every computed path is represented by a sequence of $S$ sampled positions $\mathbf{p}_j$, and the position of each sphere surrounding the omnicopter is correspondingly denoted by $\mathbf{p}_{j,i} \forall j\in\{1,...,S\} \,\, i\in \{1,...,10\}$. Consequently, a path is considered invalid if: 
\begin{equation}
\label{eq:threshold}
\exists\, j,i :
\min_{\mathbf{o}_k \in \mathcal{O}_{\text{obs}}}
\|\mathbf{p}_{j,i} - \mathbf{o}_k\|
< r_{\text{inf}} .
\end{equation}

If the trajectory is deemed as collision-inducing at any given time step, the trajectory is invalidated and discarded.

The simulated omnicopter is equipped with a depth camera~\cite{gazebo_ros_depth_camera}, which produces raw point cloud measurements. Any point clouds within the range of interest are voxelised and added to a set $\mathcal{O}_{\text{u}}$ containing the locations of all newly discovered obstacles, referred to as \emph{unknown obstacles}.
Finally, k-dimensional trees (KD-Trees) from the Point Cloud Library (PCL \cite{PCL}) are utilised to efficiently compute collisions.

\subsection{Trajectory Scoring}
\label{subsec:trajectory_scoring}
Once tentative trajectories are generated, the optimal trajectory is chosen as the one yielding the minimum value of the weighted cost function $J$, as presented in \eqref{eq:cost_function}:
\begin{align}
\label{eq:cost_function}
    J = & \, w_{\text{goal}} C_{\text{goal}} + w_{\text{path}} C_{\text{path}} + w_{\text{head}} C_{\text{head}} \nonumber \\
        & + w_{\text{look}} C_{\text{look}} + w_{\text{clear}} C_{\text{clear}} + w_{\text{face}} C_{\text{face}}.
\end{align}
where each term will be discussed below, and the corresponding weight is denoted accordingly.

The first term targets the local goal $\mathbf{p}_{\text{goal}}$, selected along the global path at a fixed lookahead offset from the current vehicle position. The associated cost, $C_{\text{goal}}$, penalises the Euclidean distance between the predicted world-frame end-position $\mathbf{p}_{S}$ and the local goal position $\mathbf{p}_{\text{goal}}$. The goal cost is defined as
\begin{equation}
\label{eq:cost_goal}
C_{\text{goal}} = \|\mathbf{p}_{S} - \mathbf{p}_{\text{goal}}\|.
\end{equation}

The path-distance cost term $C_{\text{path}}$ represents the cross-track error, defined as the minimum perpendicular distance between the predicted positions and the discretised global path. Consecutive waypoints define a set of path segments $\{\mathcal{G}_i\}$. For each segment, $\mathbf{p}_{j}$ is projected onto the corresponding line segment denoted $\mathcal{G}_j$, and the perpendicular distance is computed. The path cost is defined below:
\begin{equation}
\label{eq:cost_distance}
    C_{\text{path}} = \min_{j \in \{1,...,S\}} \left( \text{dist}(\mathbf{p}_{j}, \mathcal{G}_j) \right).
\end{equation}

The heading cost $C_{\text{head}}$ aims to penalise the orientation error between the simulated and local goal poses. The expression $\mathbf{q}_{S} \cdot \mathbf{q}_{\text{goal}}$ denotes the dot product of the simulated and desired local goal orientation unit quaternions. The absolute value accounts for the double cover property of quaternions, ensuring the shortest angular distance is computed \cite{quaternions}:
\begin{equation}
\label{eq:heading_distance}
    C_{\text{head}} = 2 \arccos\left( \left| \mathbf{q}_{S} \cdot \mathbf{q}_{\text{goal}} \right| \right).
\end{equation}
The Eigen3 library~\cite{eigen} is utilised for optimised matrix and quaternion operations.

The subsequent metric is a soft-constraint measure implemented due to the single front-facing depth camera configuration. To obtain sufficient sensing information, the vehicle should remain oriented toward the intended path segment. A lookahead point $\mathbf{p}_{\text{look}}$ is identified further along the global path via a waypoint index offset. The camera pointing direction $\hat{\mathbf{n}}_{\text{fwd}}$ is calculated by rotating the vehicle's camera axis into the world frame using the simulated orientation $\mathbf{q}_{S}$, while $\hat{\mathbf{t}}_{\text{path}}$ is defined as the unit vector from $\mathbf{p}_{S}$ toward $\mathbf{p}_{\text{look}}$. The cost $C_{\text{look}}$ is derived from the alignment of these vectors, where perfect alignment results in the camera pointing towards $\mathbf{p}_{\text{look}}$:
\begin{equation}
\label{eq:cost_lookahead}
    C_{\text{look}} = 1.0 - (\hat{\mathbf{n}}_{\text{fwd}} \cdot \hat{\mathbf{t}}_{\text{path}}).
\end{equation}

Context-aware weight alteration is implemented to facilitate path diversion around unknown obstacles. The detection of an unknown obstacle triggers a transition from Standard Mode to Agile Mode weights, activating two additional metrics. The first is the obstacle clearance cost $C_{\text{clear}}$, defined as the sum of inverse-square distances to all $k$ unknown obstacle points $\mathbf{o}_k \in \mathcal{O}_\text{u}$ within a radius $r_{\text{field}}$, as presented in \eqref{eq:unified_clearance}. This method is inspired by Artificial Potential Fields algorithms \cite{artificial_potential}:
\begin{equation}
\label{eq:unified_clearance}
    C_{\text{clear}} = \sum_{\mathbf{o}_k \in \mathcal{O}_{\text{u}}} 
    \begin{cases} 
      \frac{1}{\| \mathbf{p}_{S} - \mathbf{o}_k \|^2} & \text{if } \| \mathbf{p}_{S} - \mathbf{o}_k \| < r_{\text{field}} \\
      0 & \text{otherwise.}
    \end{cases}
\end{equation}

The second metric encourages the vehicle to orient its forward-facing camera toward nearby unknown obstacles and is denoted by $C_{\text{face}}$, where the former is denoted by the unit vector $\hat{\mathbf{n}}_{\text{fwd,face}}$ and the latter by the unit vector $\hat{\mathbf{v}}_{\text{obs,face}}$. Similar to $C_{\text{look}}$, this cost is computed using the nearest unknown obstacle point $\mathbf{o}_{\min}$. The relative obstacle direction vector is defined as:
\begin{equation}
\label{eq:cost_face}
C_{\text{face}} = 1.0 - (\hat{\mathbf{n}}_{\text{fwd,face}} \cdot \hat{\mathbf{v}}_{\text{obs,face}}),
\end{equation}

In Standard Mode, $w_{\text{clear}}$ and $w_{\text{face}}$ are set to zero. In Agile Mode, weights are adjusted to prioritise clearance over strict path following. The cost function parameters thus consist of two distinct weight sets, $\mathbf{w}_{\text{std}}$ and $\mathbf{w}_{\text{agile}}$, selected based on the presence of unknown obstacles:
\begin{equation}
\label{eq:agile_weights}
    J = 
    \begin{cases} 
      J(\mathbf{w}_{\text{agile}}) & \text{if } |\mathcal{O}_{\text{u}}| > 0 \\
      J(\mathbf{w}_{\text{std}}) & \text{if } |\mathcal{O}_{\text{u}}| = 0
    \end{cases} .
\end{equation}
where $\mathbf{w}_{\text{std}} = [w_{\text{goal}}, w_{\text{path}}, w_{\text{head}}, w_{\text{look}}, 0, 0]^\top$. Weight tuning was performed empirically. The scoring function relies on the relative weight ratios rather than absolute values. Additional metrics, such as velocity consistency, are possible but deemed not relevant for the specific goals of this implementation.

\subsection{Moving Obstacle Evasion}
\label{fast_evasion}

Fast-moving obstacles cannot be handled by the previously described 6D-DWA pipeline due to its sampling-based nature and limited reactive bandwidth. Therefore, a dedicated high-frequency subsystem is implemented for clustering, tracking, and state estimation using Kalman filtering~\cite{kalman}. Detected threats are communicated to the planner via messages containing the estimated obstacle position, velocity, and timestamp. For experimental validation, these threat messages were artificially generated to evaluate the evasion logic independently of the perception stack, while the perception task will be developed in future work.

Upon detection of a potential collision, a separate evasion pipeline is activated, as the implemented 6D-DWA is not designed for rapid avoidance of dynamic obstacles. The evasion module must generate a fast reactive manoeuvre while still respecting static-environment constraints. To maximise architectural reuse, the evasion logic leverages the same environment representation and trajectory propagation structures developed for 6D-DWA.

The decision to initiate an evasion manoeuvre is based on the predicted time to closest point of approach (CPA), denoted $t_{\text{cpa}}$, which accounts for communication latency between perception and planning nodes. Premature evasion is avoided by comparing $t_{\text{cpa}}$ to a predefined threshold. The corresponding minimum predicted separation distance between the obstacle and the omnicopter is computed as
\begin{equation}
\label{eq:d_miss}
d_{\text{miss}} = \| \mathbf{r} + \mathbf{v}_{\text{rel}} \cdot t_{\text{cpa}} \|
\end{equation}
where $\mathbf{r}$ is the relative position vector and $\mathbf{v}_{\text{rel}}$ is the relative velocity between the obstacle and the vehicle.

A collision risk is declared if 
\begin{equation}
    d_{\text{miss}} < r_{\text{bounding-sphere}} + r_{\text{obstacle}}
\end{equation}
where $r_{\text{bounding-sphere}}$ encloses the entire omnicopter and $r_{\text{obstacle}}$ represents the obstacle size estimate.

When a collision threat is confirmed, an initial evasion direction is computed as
\begin{equation}
\hat{\mathbf{e}}_{\text{opt}} =
\frac{\mathbf{v}_{\text{msg}} \times \mathbf{r}}
{\|\mathbf{v}_{\text{msg}} \times \mathbf{r}\|},
\end{equation}
which yields a unit vector orthogonal to the relative motion plane. To increase robustness, a set of candidate evasion directions $\mathcal{E}$ is generated by rotating $\hat{\mathbf{e}}_{\text{opt}}$ around the obstacle velocity unit vector $\hat{\mathbf{v}}_{\text{msg}}$.

Each candidate is forward-simulated and checked against the static map via the 10-sphere approximation. The first collision-free candidate is selected for execution. If no candidate yields a valid trajectory, the evasion manoeuvre is aborted. Upon completion of the evasion manoeuvre, the system transitions back to the 6D-DWA framework to resume nominal path following. 



\section{Experimental Results}
\label{sec:experimental_results}
This section evaluates the proposed 6D-DWA framework in simulation with respect to real-time feasibility, tracking performance, and obstacle avoidance capability. The 6D-DWA was evaluated within a ROS Noetic 
on a mobile workstation equipped with an Intel Core i7 CPU and 16\,GB of RAM. 
Experiments were conducted in Gazebo to assess computational scaling, the effectiveness of adaptive velocity sampling, sensitivity to scoring weights, static obstacle avoidance behaviour, Agile Mode performance in unknown obstacle scenarios, and the fast evasion pipeline for dynamic threats. 
\subsection{Computational Performance}
\label{subsec:computational_performance}


To verify real-time feasibility, we measured the computation time per 6D-DWA planning iteration, $t_{\text{loop}}$, under variations in (i) the number of sampled velocity commands $S$ and (ii) the number of voxelised static obstacle points in the local map $N_o$. The planner targets a 5~Hz replanning rate, corresponding to a maximum allowable loop time of $t_{\max}=0.2$~s.

\begin{figure}[t]
    \centering
    \includegraphics[width=\columnwidth, trim=0 0 0 0cm, clip]{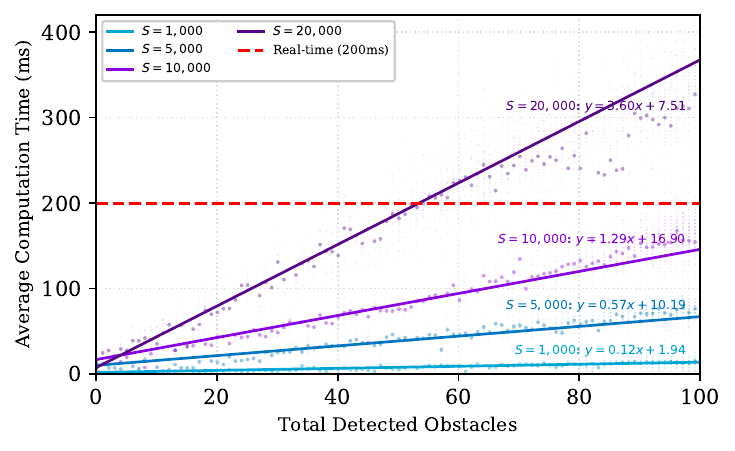}
    \caption{Average computation time vs. number of obstacles with linear regression lines. Increased sample counts $S$ lead to exceeding the real-time threshold at lower obstacle densities.}
    \label{fig:test_comp_3}
\end{figure}

Fig.~1 shows that $t_{\text{loop}}$ increases with both $S$ and $N_o$. For each fixed sampling configuration, we fit a linear model of the form $t_{\text{loop}} \approx a N_o + b$ to the corresponding cluster of measurements in Fig.~1. The fitted lines are used to summarise the trend and to estimate the obstacle density at which the runtime budget is violated. For example, for $S=20{,}000$, the fitted model shows that the average loop time exceeds $0.2$~s when more than approximately 54 obstacle points are present in the local map, resulting in deteriorated tracking error.

These results highlight an operational trade-off: increasing $S$ improves planning quality but reduces the allowable obstacle density under the fixed time budget, while increasing the local-map radius or reducing the voxel size increases $N_o$ and therefore the runtime. Consequently, sampling density should be increased only when the resulting $t_{\text{loop}}$ remains consistently below $t_{\max}$.

\subsection{Velocity Sampling Evaluation}
\label{sampling_testing}
\begin{figure}[t]
    \centering
    \includegraphics[width=\columnwidth, trim=0.2 2.2cm 0 0.8cm, clip]{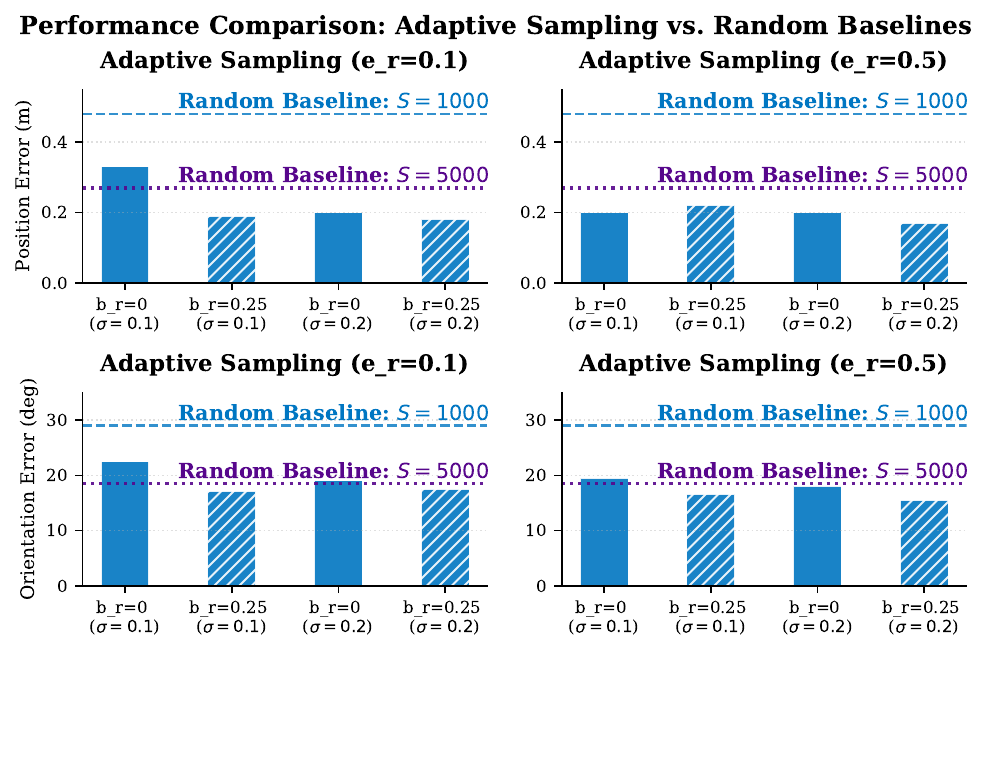}
    \caption{Average cross-track and orientation errors for various sampling configurations. Adaptive sampling significantly outperforms random sampling.}
    \label{fig:sampling_1}
\end{figure}

The adaptive sampling evaluation outcomes are presented in Fig.~\ref{fig:sampling_1}. As expected, random sampling showed lower errors for a higher number of samples. Notably, all configurations with adaptive sampling performed better in both metrics than the purely random 1,000-sample configuration. The implementation of adaptive sampling augmented the performance of the planner given a set number of velocity samples. Due to the high dimensionality of the 6D velocity search space, random sampling was less likely to yield well-performing velocity samples. The impact of adaptive sampling should decrease for lower velocity limits or higher numbers of samples.

Diminished performance for an Exploration ratio ($e_r$) of $0.1$, a Boundary-Search ratio ($b_r$) of $0$, and a Focused-Search sampling deviation ($\sigma$) of $0.1$ was attributed to insufficient exploration of the velocity search space. As $90\%$ of the samples were focused closely around the previous best solution, the exploration of other velocities was limited. The variations of the Focused-Search ratio ($f_r$) and $b_r$ showed no clear trend. The configurations with $e_r = 0.5$ and $\sigma = 0.1$ consistently exhibited performance on-par with or superior to the random 5,000-sample configuration. Consequently, an $e_r$ of $0.5$, $f_r$ of $0.25$, $\sigma$ of $0.1$, and $b_r$ of $0.25$ were used in further testing.

\subsection{6D-DWA Weight Tuning}
\label{weight_testing}
To identify the optimal 6D-DWA weight distribution, a grid search was performed by varying the weights of the core 6D-DWA scoring function in a curved global path-following task. Notably, the percentage of the total weight sum determines each 6D-DWA scoring metric’s impact on omnicopter behaviour, rather than the weight's absolute value.

\begin{figure}[b]
    \centering
    \includegraphics[width=\columnwidth, trim=0.2 0.2cm 0 1.2cm, clip]{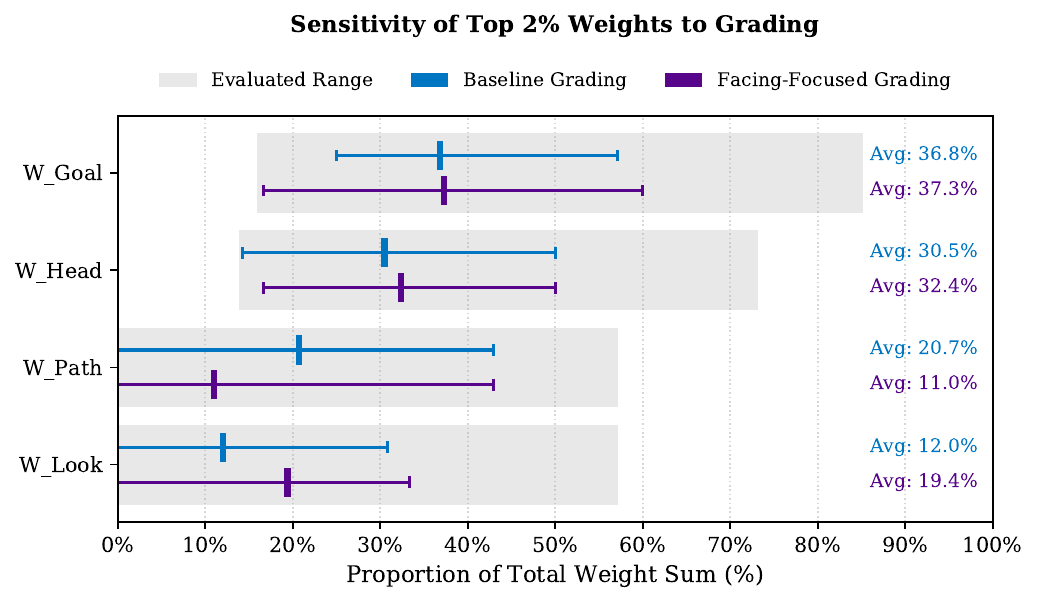}
    \caption{Averaged best performing weight configurations. Desired behaviour definition via a trial-grading function affects the optimal 6D-DWA weight distribution.}
    \label{fig:tuning_3}
\end{figure}

A weighted trial-grading function was determined based on the average cross-track error, average orientation heading error, total covered distance, and average 'lookahead' error. The top $2\%$ best-grade 6D-DWA weight configurations for each grading scenario were averaged. Two grading configurations are presented in Fig.~\ref{fig:tuning_3}: Baseline and Facing-Focused, with the latter prioritises minimising 'lookahead' error.

The superior runs across all grading scenarios resulted in a relative average $w_{goal}$ of approximately $37\%$. Similarly, the various grading configurations showed little impact on the average relative $w_{head}$, which averaged approximately $31\%$. As expected, grading scenarios with a higher contribution of the lookahead metric favoured higher $w_{look}$ values. Intuitively, a higher $w_{look}$ weight ensures the camera remains aligned with the path, which is critical for maintaining environmental awareness on a single-sensor platform.

These results indicate that weights must be tuned for the specific desired behaviour of the application. Consequently, baseline weights for further experiments were set to 40, 30, 20, and 10 for $w_{goal}$, $w_{head}$, $w_{path}$, and $w_{look}$, respectively.

\begin{figure}[t]
    \centering
    \includegraphics[width=0.85\columnwidth, trim=0.2 0.4 0 1.7cm, clip]{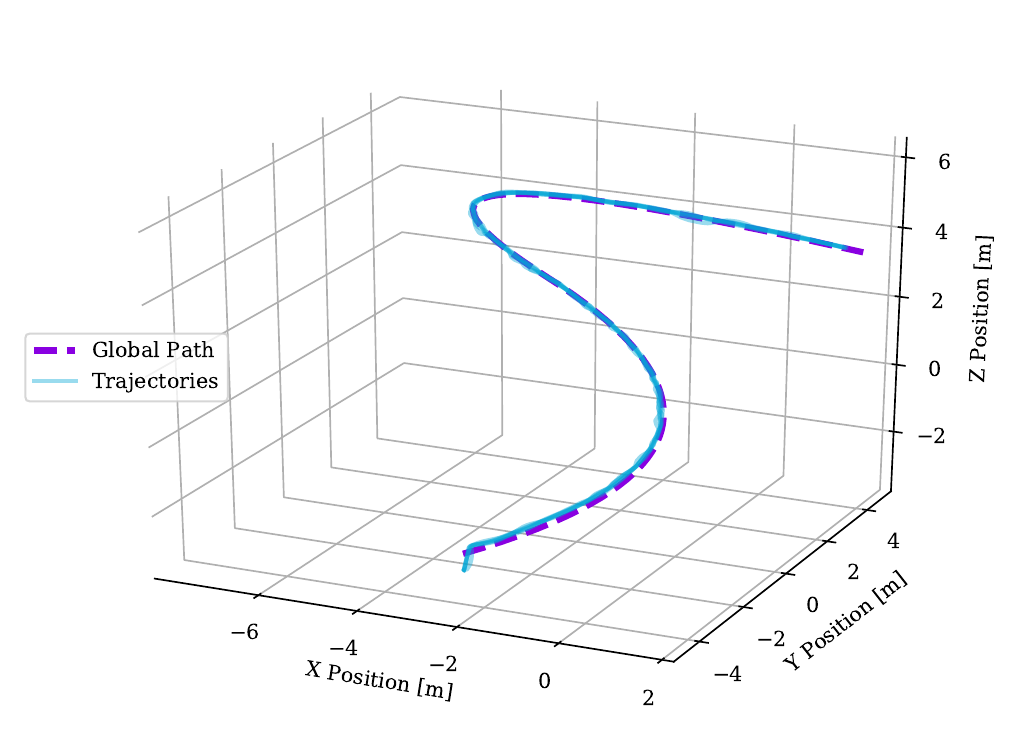}
    \caption{Resultant trajectories for dense-waypoint global path following. The 6D-DWA successfully commands the omnicopter through the empty environment.}
    \label{fig:test_path_1}
\end{figure}

The path-following performance of the 6D-DWA given a waypoint-dense global path plan, such as those generated by the RRT* algorithm, was assessed.  As seen in Fig.~\ref{fig:test_path_1}, the omnicopter accurately followed the global path. The system maintained an average distance to the path of under $0.1~\text{m}$. The average orientation error was approximately $13^{\circ}$, while the average lookahead error was $10^{\circ}$. Notably, the orientation and lookahead directions can differ depending on the path geometry, waypoint poses, vehicle position, and the selected waypoint horizons.



\subsection{Static Obstacle Avoidance Evaluation}
\label{subsec:static_obstacle_avoidance}
The static obstacle experiments evaluate the planner’s ability to (i) reject infeasible segments of the global path, (ii) generate collision-free detours, and (iii) traverse geometrically constrained passages while balancing orientation and global path-following objectives.
\begin{figure}[t]
    \centering
    \includegraphics[width=0.98\columnwidth, trim=0.2 0.2 0 0.6cm, clip]{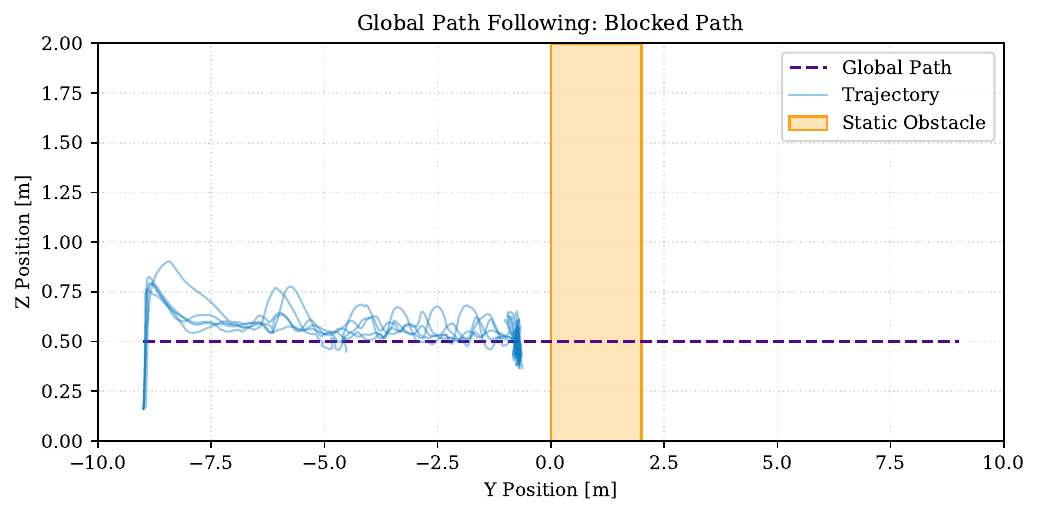}
    \caption{Side view of the trajectories from following a blocked global path. The 6D-DWA correctly recognises the obstacle and avoids collision.}
    \label{fig:static_1}
\end{figure}

In the first scenario, a wall defined in the STL map directly obstructed the global path. The obstacles were not assigned physical collision properties in the Gazebo world; therefore, avoidance behaviour relied solely on the planner’s internal collision-checking mechanism. As shown in Fig.~\ref{fig:static_1}, the 6D-DWA did not command any trajectory intersecting the wall, confirming the reliability of the voxelised environment representation and sphere-based vehicle approximation.

\begin{figure}[b]
    \centering
    \includegraphics[width=0.98\columnwidth, trim=0.2 0.2 0 0.6cm, clip]{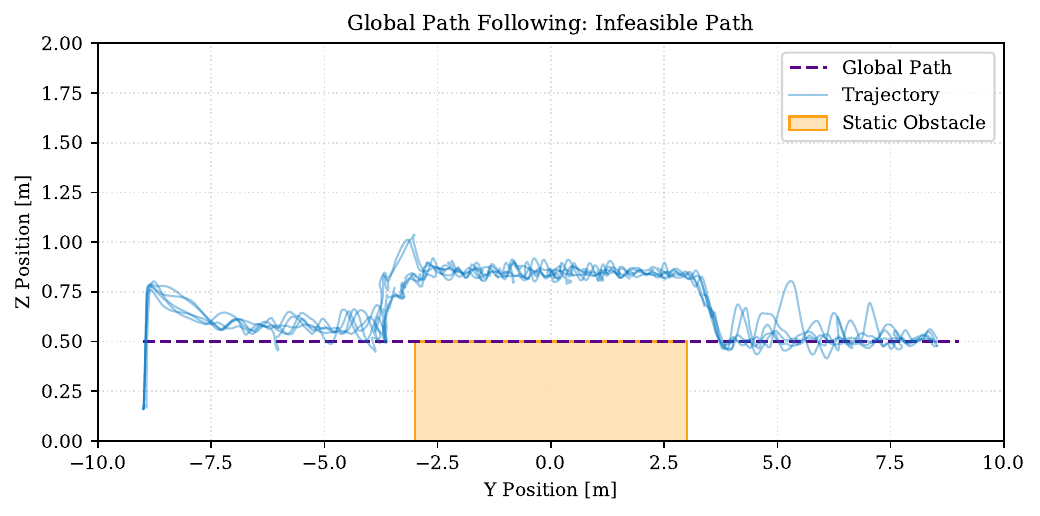}
    \caption{Side view of the trajectories from following an infeasible global path. The 6D-DWA correctly identifies the obstacle, detours, and rejoins the global path when feasible.}
    \label{fig:static_2}
\end{figure}

In the second scenario, a known static obstacle required deviation from the global path, followed by rejoining once feasible. As illustrated in Fig.~\ref{fig:static_2}, the planner correctly identified the infeasibility of the obstructed segment, generated a detour maintaining safe clearance, and subsequently converged back to the global path after clearing the obstacle. This demonstrates that the weighted objective allows temporary relaxation of strict path adherence while preserving long-term global progress.

\begin{figure}[t]
    \centering
    \includegraphics[width=0.85\columnwidth, trim=1.9cm 0.2 0 2.2cm, clip]{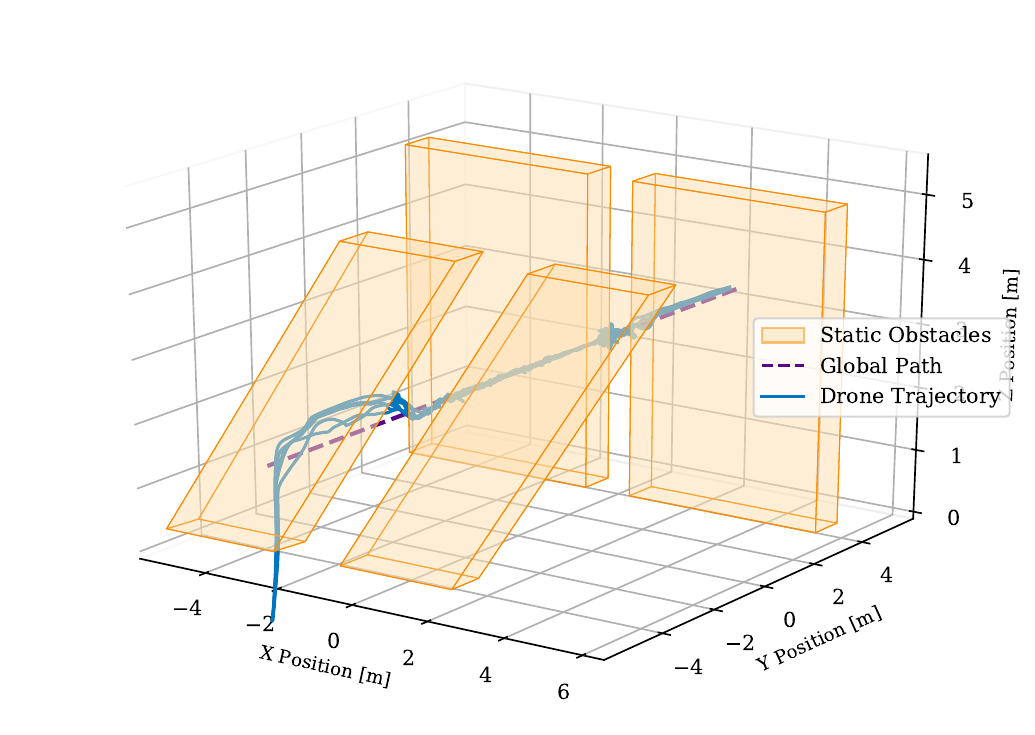}
    \caption{Path following of the orientation-infeasible global path. The 6D-DWA manages to alter the orientation to fit through the STL-defined gaps.}
    \label{fig:gaps_traj123}
\end{figure}
The third scenario investigated traversal through narrow gaps whose orientations were misaligned with the global path direction shown in Fig.~\ref{fig:gaps_traj123}. Two gaps of 1.3\,m width were defined in the STL map: one rotated by $45^\circ$ and the other by $90^\circ$ relative to the local tangent of the global path. All tested configurations successfully traversed the $45^\circ$ gap. However, configurations with the baseline heading weight $w_{\text{head}}=30$ failed to pass the $90^\circ$ gap, instead halting before the opening. Reducing $w_{\text{head}}$ enabled successful traversal by permitting larger orientation deviations from the global path.


This result highlights a critical trade-off: excessive emphasis on orientation consistency can artificially restrict feasible motion in geometrically constrained environments. Lowering $w_{\text{head}}$ relaxes this constraint, permitting the vehicle to adopt orientations necessary for traversal through misaligned structures. The experiment demonstrates that orientation weighting directly influences geometric feasibility in 6D planning.

During gap traversal, occasional proximity of the sphere centres to obstacle surfaces below the commanded safety margin was observed. This behaviour is attributed to controller overshoot and tracking error rather than deficiencies in the collision-checking formulation. These results highlight the importance of the radius inflation used in collision checking to accommodate for controller tracking errors.

\subsection{Agile Mode Weight Tuning}
\label{subsec:agile_weight_tuning}

The Agile Mode experiments evaluate the planner’s ability to divert from the global path when previously unmapped obstacles are detected. Upon identification of unknown obstacle points, the planner switches from Standard Mode weights to Agile Mode weights, prioritising obstacle clearance over strict adherence to the global path.

An unknown cuboid obstacle was introduced beyond the detection threshold of any pre-mapped static obstacle, ensuring it was classified as an unknown threat. Two configurations were tested: (i) an off-centre obstacle located adjacent to the global path, and (ii) a centred obstacle directly intersecting the global path. In Agile Mode, $w_{\text{path}}$ was set to zero to allow unrestricted deviation from the global path during obstacle avoidance, while other weights were modified to accommodate for the current mode.

\begin{figure}[t]
    \centering
    \includegraphics[width=0.97\columnwidth, trim=0 0.2cm 0 0.2cm, clip]{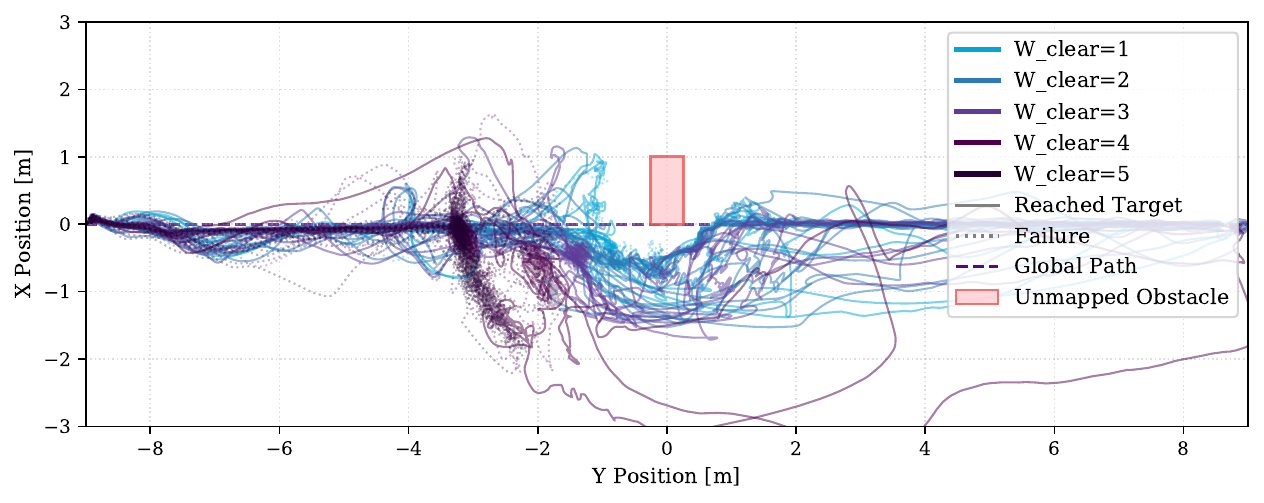}
    \caption{Impact of unknown obstacle clearance weight $w_{clear}^{agile}$ on off-centre trajectories. The higher the clearance weight the lower the allowable obstacle proximity.}
    \label{fig:test_obst}

    \vspace{0.2cm} 

    \includegraphics[width=0.97\columnwidth, trim=0 0.2cm 0 0.2cm, clip]{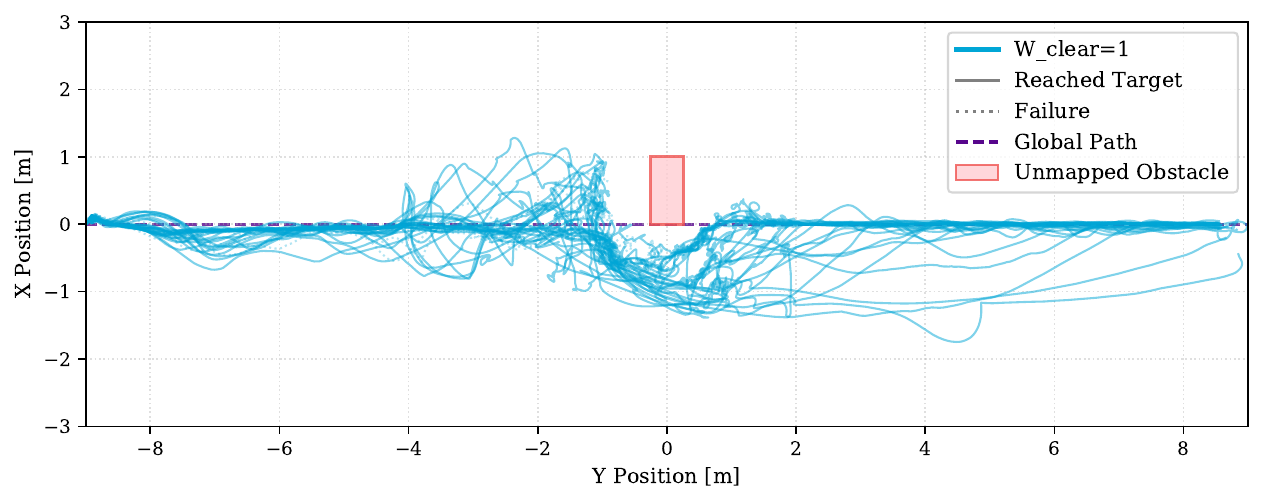}
    \caption{Trajectories for $w_{clear}^{agile}=1.0$ with an off-centre placed obstacle. The 6D-DWA manages to command the omnicopter around the obstacle in $79.3\%$ of the trials.}
    \label{fig:test_obst_3}
\end{figure}

\paragraph{Off-centre obstacle scenario}
The resulting trajectories for the off-centre obstacle scenario are plotted in Fig.~\ref{fig:test_obst}. The figure shows that the planner consistently detected the unknown obstacle and activated Agile Mode. The clearance weight $w_{\text{clear}}$ was varied to assess its influence on avoidance behaviour. For $w_{\text{clear}} = 1.0$, the planner achieved a success rate of 72.2\%, increasing to 79.3\% when combined with $w_{\text{head}} = 0.0$ as illustrated in Fig.~\ref{fig:test_obst_3}. Increasing $w_{\text{clear}}$ beyond this value reduced success rates, reaching 0\% at $w_{\text{clear}} = 5.0$.
Higher clearance weights produced stronger repulsive behaviour, generating trajectories that maintained larger separation from the obstacle. However, excessive repulsion prevented sufficient forward progress along the global path, causing the planner to stall or enter a timeout condition. This demonstrates the need to balance clearance with goal-seeking behaviour to avoid artificial deadlock.

\begin{figure}[t]
    \centering
    \includegraphics[width=0.97\columnwidth, trim=0 0.2cm 0 0.2cm, clip]{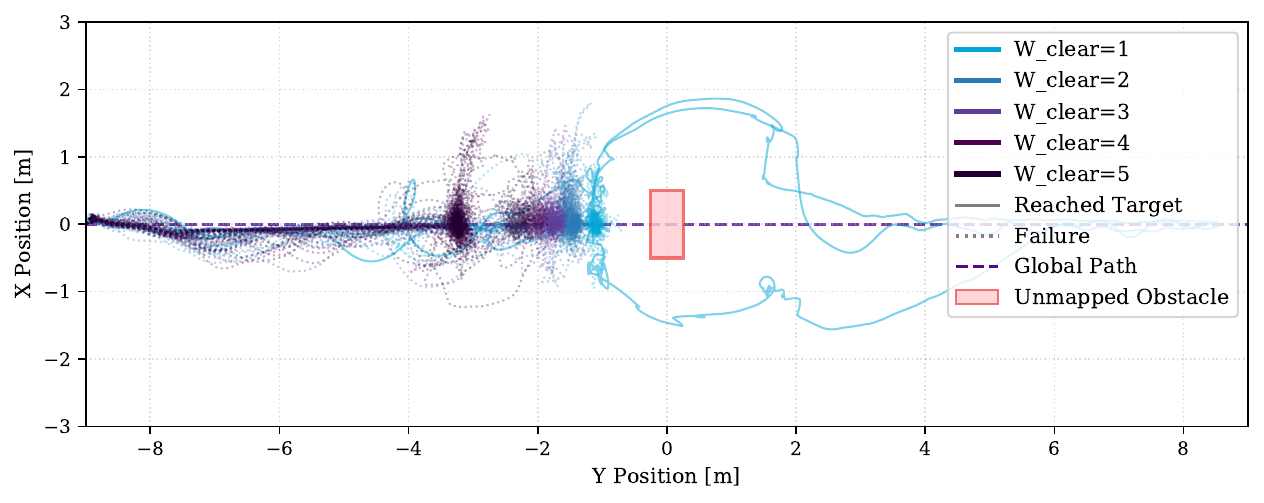}
    \caption{Effect of $w_{clear}^{agile}$ on a centrally placed unknown obstacle trajectory. High clearance weights forbid exploration close to the obstacle.}
    \label{fig:test_obst_2}

    \vspace{0.2cm} 

    \includegraphics[width=0.98\columnwidth, trim=0 0.2cm 0 0.2cm, clip]{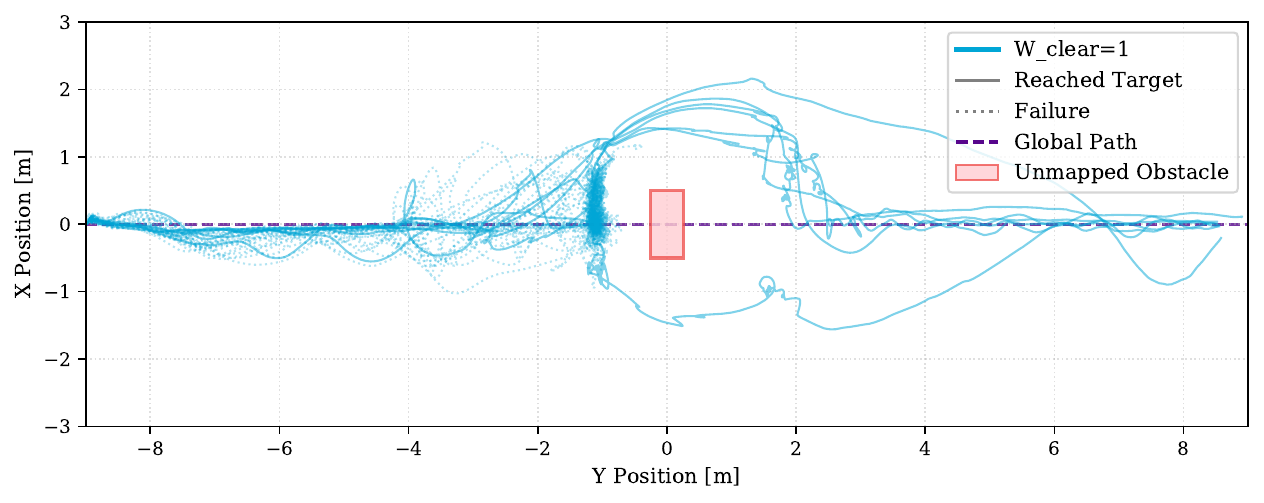}
    \caption{Trajectories for $w_{clear}^{agile}=1.0$ with a centred unknown obstacle. The 6D-DWA manages to command the omnicopter around the obstacle in $41.4\%$ of the trials.}
    \label{fig:test_obst_4}
\end{figure}

\paragraph{Centred obstacle scenario}
The centred obstacle configuration, shown in Fig.~\ref{fig:test_obst_2}, was significantly more challenging. For $w_{\text{clear}} = 1.0$ and $w_{\text{head}} = 0.0$, the success rate was 41.4\% as illustrated in Fig.~\ref{fig:test_obst_4}. Larger clearance weights consistently resulted in failure. In this symmetric configuration, the planner must commit to a lateral escape direction without strong directional bias. High repulsive forces combined with limited exploration capability led to local entrapment, illustrating the inherent local-minima limitations of reactive, short-horizon planners.

Interestingly, successful avoidance trajectories predominantly diverted toward positive $x$ coordinates, a bias attributed to the initial vehicle orientation. The result highlights the sensitivity of local reactive planners to initial conditions in symmetric environments.

\paragraph{Effect of facing weight}
The obstacle-facing weight $w_{\text{face}}$ did not significantly affect collision avoidance success. A moderate value (e.g., $w_{\text{face}} = 10$) was sufficient to maintain camera alignment with the obstacle during diversion. However, higher values delayed the transition back to Standard Mode by maintaining focus on the obstacle even after safe clearance was achieved. Lower $w_{\text{face}}$ values facilitated faster re-alignment with the global path following obstacle traversal.

Overall, the Agile Mode experiments demonstrate that adaptive weight switching enables effective diversion from the global path in the presence of unknown obstacles. However, the success rate strongly depends on balancing clearance and orientation constraints, and performance degrades in symmetric or highly constrained configurations due to the limited foresight of short-horizon local planning.

\subsection{Evasion Evaluation}
\label{subsec:evasion_testing}

The evasion experiments assess the effectiveness of the dedicated fast-reaction pipeline in handling dynamic collision threats that cannot be addressed within the standard 6D-DWA replanning cycle. Evasion is triggered by threat messages containing estimated obstacle position and velocity, from which the time to closest point of approach (CPA) and minimum predicted separation distance are computed.

\begin{figure}[b]
    \centering
    \includegraphics[width=0.95\columnwidth, trim=0 0.4cm 0 1.7cm, clip]{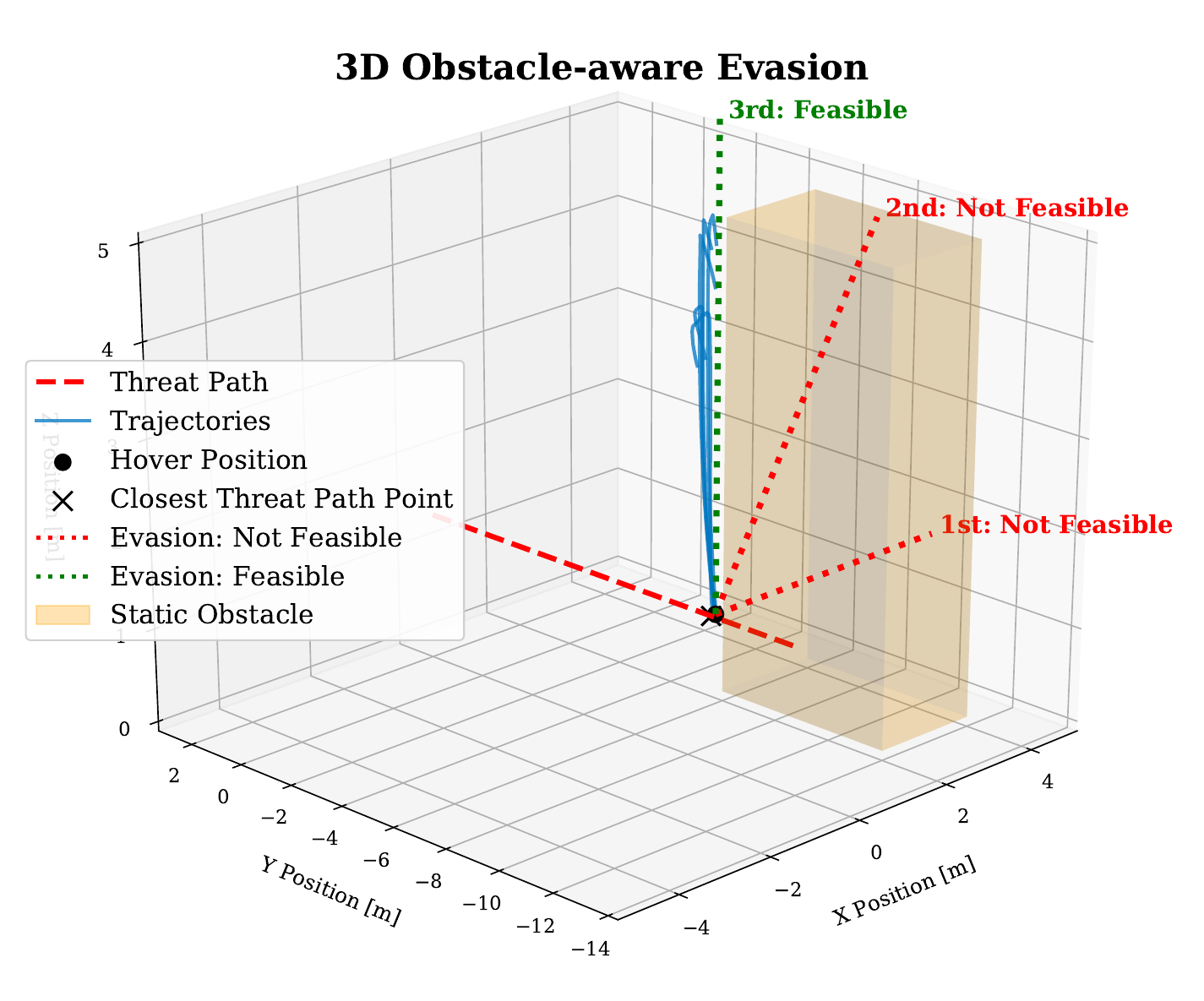}
    \caption{Evasion manoeuvre in the presence of static obstacles. The evasion logic correctly determines evasion directions and their feasibility.}
    \label{fig:test_evasion_2}
\end{figure}

The evasion was evaluated in the presence of static obstacles. Simulated threat messages were constructed to generate an initially infeasible evasion direction, as shown in Fig.~\ref{fig:test_evasion_2}. As shown in the figure, the first candidate evasion vector was blocked by static geometry, requiring evaluation of subsequent rotated candidate vectors. In all trials, the evasion module successfully identified a collision-free direction and executed a safe manoeuvre. This confirms that the candidate-generation strategy and collision validation against the static map are sufficient to resolve conflicts between dynamic and static constraints.

The evasion prioritisation was tested during a global path-following task. Upon detection of an impending collision, the system transitioned from 6D-DWA to the fast evasion pipeline. All trials resulted in successful avoidance, after which the state machine returned control to the 6D-DWA planner and resumed global path tracking. This demonstrates correct prioritisation of reactive safety behaviour over nominal path-following and validates the state-machine architecture for pipeline switching.



\section{Conclusion}
\label{conclusion}
This work presented a real-time 6D extension of the Dynamic Window Approach for an omnidirectional multirotor platform. Through adaptive velocity sampling, sphere-based geometry approximation, and voxelised local mapping, the planner operates within a 5~Hz replanning cycle while maintaining accurate tracking of waypoint-dense global paths. 

Experimental results demonstrate reliable static obstacle avoidance, effective diversion from unknown obstacles via context-aware weight adaptation, and successful integration of a fast evasion pipeline for dynamic threats. The experiments further highlight the sensitivity of 6D planning to weight selection and the inherent local-minima limitations of short-horizon reactive methods in highly constrained or symmetric environments.

Future work will focus on improving robustness in multi-obstacle dynamic scenarios, incorporating uncertainty-aware safety margins, and validating the framework on physical hardware.

\section*{Acknowledgment}

This research was carried out in collaboration between Delft University of Technology and New York University Abu Dhabi. The work was partially supported through the Center for Artificial Intelligence and Robotics, funded by Tamkeen under grant CG010.
\bibliographystyle{IEEEtran}
\bibliography{references}

@article{UAVpath,
  title={UAV path planning techniques: A survey},
  author={Ghambari, Soheila and Golabi, Mahmoud and Jourdan, Laetitia and Lepagnot, Julien and Idoumghar, Lhassane},
  journal={RAIRO-Operations Research},
  volume={58},
  number={4},
  pages={2951--2989},
  year={2024},
  publisher={EDP Sciences}
}

@article{ClassificationUAV,
  title={Applications and classifications of unmanned aerial vehicles: A literature review with focus on multi-rotors},
  author={Sabour, MH and Jafary, P and Nematiyan, S},
  journal={The Aeronautical Journal},
  volume={127},
  number={1309},
  pages={466--490},
  year={2023},
  publisher={Cambridge University Press}
}

@article{Fully,
  title={Fully actuated multirotor UAVs: A literature review},
  author={Rashad, Ramy and Goerres, Jelmer and Aarts, Ronald and Engelen, Johan BC and Stramigioli, Stefano},
  journal={IEEE Robotics \& Automation Magazine},
  volume={27},
  number={3},
  pages={97--107},
  year={2020},
  publisher={IEEE}
}

@article{aerialmanipulation,
  title={Aerial manipulation: A literature review},
  author={Ruggiero, Fabio and Lippiello, Vincenzo and Ollero, Anibal},
  journal={IEEE Robotics and Automation Letters},
  volume={3},
  number={3},
  pages={1957--1964},
  year={2018},
  publisher={IEEE}
}

@article{local1,
  title={A review of UAV path-planning algorithms and obstacle avoidance methods for remote sensing applications},
  author={Debnath, Dipraj and Vanegas, Fernando and Sandino, Juan and Hawary, Ahmad Faizul and Gonzalez, Felipe},
  journal={Remote Sensing},
  volume={16},
  number={21},
  pages={4019},
  year={2024},
  publisher={MDPI}
}

@article{pathsurvey,
  title={UAV Path Planning Using Optimization Approaches: A Survey: AA Saadi et al.},
  author={Ait Saadi, Amylia and Soukane, Assia and Meraihi, Yassine and Benmessaoud Gabis, Asma and Mirjalili, Seyedali and Ramdane-Cherif, Amar},
  journal={Archives of Computational Methods in Engineering},
  volume={29},
  number={6},
  pages={4233--4284},
  year={2022},
  publisher={Springer}
}

@article{meta,
  title={Path planning optimization in unmanned aerial vehicles using meta-heuristic algorithms: A systematic review},
  author={Yahia, Hazha Saeed and Mohammed, Amin Salih},
  journal={Environmental Monitoring and Assessment},
  volume={195},
  number={1},
  pages={30},
  year={2023},
  publisher={Springer}
}

@INPROCEEDINGS{nyuad_drone,
  author={Hamandi, Mahmoud and Ali, Abdullah Mohamed and Kyriakopoulos, Konstantinos and Tzes, Anthony and Khorrami, Farshad},
  booktitle={2025 IEEE International Conference on Robotics and Automation (ICRA)}, 
  title={An Omnidirectional Non-Tethered Aerial Prototype with Fixed Uni-Directional Thrusters}, 
  year={2025},
  volume={},
  number={},
  pages={8649-8655},
  doi={10.1109/ICRA55743.2025.11128060}}

@INPROCEEDINGS{nyuad_RRT*,
  author={Ali, Abdullah Mohamed and Hamandi, Mahmoud and Tzes, Anthony},
  booktitle={2025 International Conference on Unmanned Aircraft Systems (ICUAS)}, 
  title={Efficient Safe Trajectory Planning for an Omnidirectional Drone}, 
  year={2025},
  volume={},
  number={},
  pages={785-792},
  doi={10.1109/ICUAS65942.2025.11007909}}

@ARTICLE{dwa1,
  author={Fox, D. and Burgard, W. and Thrun, S.},
  journal={IEEE Robotics \& Automation Magazine}, 
  title={The dynamic window approach to collision avoidance}, 
  year={1997},
  volume={4},
  number={1},
  pages={23-33},
  doi={10.1109/100.580977}}

@misc{gazebo_ros_depth_camera,
  title        = {ROS Depth Camera Integration},
  author       = {{Gazebo Tutorials}},
  year         = {2014},
  howpublished = {\url{https://classic.gazebosim.org/tutorials}},
  note         = {Accessed: Dec. 21, 2025}
}

@inproceedings{PCL,
  title={3d is here: Point cloud library (pcl)},
  author={Rusu, Radu Bogdan and Cousins, Steve},
  booktitle={2011 IEEE international conference on robotics and automation},
  pages={1--4},
  year={2011},
  organization={IEEE}
}

@article{quaternions,
  title={Quaternion kinematics for the error-state Kalman filter},
  author={Sola, Joan},
  journal={arXiv preprint arXiv:1711.02508},
  year={2017}
}

@inproceedings{artificial_potential,
  title={Real-time obstacle avoidance for manipulators and mobile robots},
  author={Khatib, Oussama},
  booktitle={Proceedings. 1985 IEEE international conference on robotics and automation},
  volume={2},
  pages={500--505},
  year={1985},
  organization={IEEE}
}

@article{3d-dwa,
  title={DWA-3D: A reactive planner for robust and efficient autonomous UAV navigation in confined environments},
  author={Bes, Jorge and Dendarieta, Juan and Riazuelo, Luis and Montano, Luis},
  journal={Robotics and Autonomous Systems},
  pages={105196},
  year={2025},
  publisher={Elsevier}
}

@article{kalman,
  title={An introduction to the kalman filter},
  author={Bishop, Gary and Welch, Greg and others},
  journal={Proc of SIGGRAPH, Course},
  volume={8},
  number={27599-23175},
  pages={41},
  year={2001}
}

@INPROCEEDINGS{omniiros,
  author={Hamandi, Mahmoud and Ali, Abdullah Mohamed and Tzes, Anthony and Khorrami, Farshad},
  booktitle={2025 IEEE/RSJ International Conference on Intelligent Robots and Systems (IROS)}, 
  title={Experimental Evaluation of Safe Trajectory Planning for an Omnidirectional UAV}, 
  year={2025},
  volume={},
  number={},
  pages={11104-11111},}

@inproceedings{gazebo,
  author = "N. Koenig and A. Howard",
  title = "Design and Use Paradigms for Gazebo, An Open-Source Multi-Robot Simulator",
  booktitle = "{IEEE/RSJ} International Conference on Intelligent Robots and Systems",
  pages = "2149-2154",
  address = "Sendai, Japan",
  month = "Sep",
  year = "2004",
}

@inproceedings{ros,
author = {Quigley, M. and Conley, K. and Gerkey, B. and Faust, J. and Foote, T. and Leibs, J. and Wheeler, B. and Ng, A.},
year = {2009},
month = {01},
pages = {},
title = {{ROS}: An Open-Source Robot Operating System},
volume = {3},
journal = {{ICRA} Workshop on Open Source Software}
}

@article{rrtstar,
author = {Nasir, J. and Islam, F. and Ayaz, Y.},
year = {2013},
month = {12},
pages = {39-51},
title = {Adaptive Rapidly-Exploring-Random-Tree-Star ({RRT*}) -Smart: Algorithm Characteristics and Behavior Analysis in Complex Environments},
volume = {02},
journal = {Asia-Pacific Journal of Information Technology and Multimedia},
doi = {10.17576/apjitm-2013-0202-04}
}

@article{voxel,
title = {Voxel-based Representation of 3D Point Clouds: Methods, Applications, and Its Potential Use in the Construction Industry},
journal = {Automation in Construction},
volume = {126},
pages = {103675},
year = {2021},
issn = {0926-5805},
doi = {https://doi.org/10.1016/j.autcon.2021.103675},
author = {Y. Xu and X. Tong and U. Stilla},
}

@misc{eigen,
  author       = {Guennebaud, G. and Jacob, B. and others},
  title        = {Eigen v3: {C++} Template Library for Linear Algebra},
  howpublished = {\url{https://libeigen.gitlab.io}},
  year         = {2010},
  note         = {Accessed: 2026-01-25}
}

@article{algcomp,
author = {Radmanesh, M. and Kumar, M. and Guentert, P. and Sarim, M.},
year = {2018},
month = {04},
pages = {1-24},
title = {Overview of Path Planning and Obstacle Avoidance Algorithms for {UAV}s: A Comparative Study},
volume = {6},
journal = {Unmanned Systems},
doi = {10.1142/S2301385018400022}
}

@misc{full_act,
      title={Integration of Fully-Actuated Multirotors into Real-World Applications}, 
      author={A. Keipour and M. Mousaei and A. T. Ashley and S. Scherer},
      year={2021},
      archivePrefix={arXiv},
      primaryClass={cs.RO},
      doi={10.48550/arXiv.2011.06666}
}

@article{Occupancygrid,
author = {Elfes, A.},
year = {1989},
month = {07},
pages = {46 - 57},
title = {Using Occupancy Grids for Mobile Robot Perception and Navigation},
volume = {22},
journal = {Computer},
doi = {10.1109/2.30720}
}

\end{document}